\documentclass[10pt,a4paper]{article}

\usepackage[T1]{fontenc}
\usepackage[latin1]{inputenc}
\usepackage[american]{babel}
\usepackage[total={16cm,25cm}]{geometry}
\usepackage{cite}
\usepackage{graphicx}
\usepackage[nottoc]{tocbibind}

\title{Solution Representations and Local Search for the\\bi-objective Inventory Routing Problem}
\author{Thibaut Barthélemy$^{1,}$\footnote{Corresponding author, {\tt barthelt@hsu-hh.de}} \hspace{3em} Martin Josef Geiger$^1$ \hspace{3em} Marc Sevaux$^{2}$\\[2ex]
{\small \parbox{.45\textwidth}{
  \begin{center}
    $^1$Helmut Schmidt University,\\
    University of the Federal Armed Forces Hamburg\\
    Logistics Management Department\\
    Hamburg, Germany
  \end{center}}
\parbox{.45\textwidth}{
  \begin{center}
    $^2$Universit\'e de Bretagne-Sud\\
    Lab-STICC\\Lorient, France
  \end{center}}
}}

\date{}

\begin{document}
\maketitle

\vspace{-0.6cm}

\section{Introduction}

The solution of the biobjective IRP is rather challenging, even for metaheuristics. We are still lacking a profound understanding of appropriate solution representations and effective neighborhood structures. Clearly, both the delivery volumes and the routing aspects of the alternatives need to be reflected in an encoding, and must be modified when searching by means of local search. Our work contributes to the better understanding of such solution representations. On the basis of an experimental investigation, the advantages and drawbacks of two encodings are studied and compared.

\section{Representations}

Due to the high complexity of the problem, many solution approaches introduce some simplifying assumptions. Some authors restrict quantities to deliver \cite{campbell04} or allow backlogging \cite{abdelmaguid09}. Cycling visiting was also studied \cite{raa09} and other authors limit routing possibilities \cite{li08}.

To reduce the combinatorial complexity, we choose a specific replenishment strategy such that shipped amounts are not part of the searched variable set anymore; moreover, visits are periodic \cite{geiger11_art}. In our first representation, only the visiting frequency $f_i$ of each customer $i$ has to be decided; shipping amounts are deduced from our replenishment strategy. In our new representation, the starting date $s_i$ of the periodic visiting sequence is adjustable; one delivery before the starting date is allowed if required to cover first consumptions of some customers.

Our search algorithms will have to explore values of $n$ or $2n$ variables, instead of up to $Hn$ variables for other approaches ($H$ is the size of the time horizon). Such simplifications make us miss solutions and in particular some non-dominated ones. Nevertheless, we expect that the loss of freedom in quantities and visiting dates is balanced by the higher local search speed.

\section{Local Search and Evaluation}

Not surprisingly, solution approaches often employ heuristics and meta-heuristics \cite{moin07}. Following in these footsteps, we began to experiment our representation with an heuristic. For the moment, our neighborhood operator is simple and we investigate representations and solution selection strategies.

The first selection strategy uses reference points \cite{geiger11_conf}. At each iteration, among all non dominated solutions, we select the $R$ closest solutions to $R$ evenly spread reference points. Our second strategy is inspired from NSGA-II \cite{deb02}. At each iteration, among all non dominated solutions, a subset of $R$ solutions is randomly chosen. The probability that a solution is taken depends on its crowding distance.

Neighbors are generated for each of the $R$ alternatives then inserted among all other solutions and dominated solutions are removed. This process is repeated until the total number of evaluations reaches a limit chosen by the user.


The local search adjusts starting dates and frequencies of the visits. From our short representations, the role of the evaluation phase is to compute holding costs and routing costs of each solution. Whereas inventory costs are fast to compute exactly with our replenishment strategy, the routing cost evaluation is a NP-Hard problem and we must restrict ourselves to a rough value.

The sum of inventory levels $L_i^t$ of all customers $i$ over all dates $t$ represents the overall holding cost: $
z_1 = \sum_{t=1}^{H} \sum_{i=1}^n L_i^t
$.
Evaluation of inventories takes $\Theta(H n)$.

Concerning routing costs, we get them from the sum of the costs of the VRP solutions for all dates:
$
z_2 = \sum_{t=1}^{H} \textrm{VRP}(q^t)
$,
where $q^t$ is the set of quantities to deliver to customers visited at the date $t$.
The route planning is preliminary solved by the Clarke \& Wright's heuristic, which allows for a fast construction of tours in O($n^2 \log n$). Then, a 2-Opt operator optimizes the tours independently in O($n^2$) in average.

\section{Experiments and Results}

We denote by $ev$ the number of solutions evaluated since the beginning of the search. It is interesting to note that the number of modifications done on any initial solution during search is at most $\frac{ev}{4nR}$ and that the average number of modifications applied per customer can not exceed $\frac{ev}{4n^2 R}$.
In our previous approach, a customer was served for the first time just before going to stockout \cite{geiger11_art}. As described above, starting dates are now movable. We noticed that this new freedom for delivery dates is always beneficial when $\frac{ev}{4n^2 R}$ is constant with regard to $n$. This is a typical result on the left:
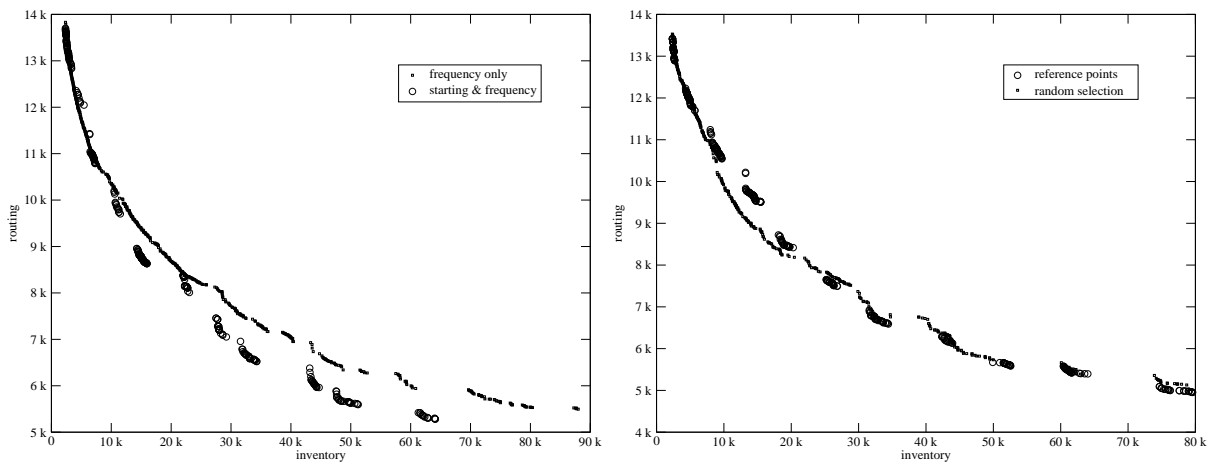
\begin{center}
    \includegraphics[width=0.49\textwidth]{single-vs-double.eps}
    \includegraphics[width=0.49\textwidth]{refpoints-vs-randomselec.eps}
\end{center}
On the right, we notice that our random selection method reduces the gaps, but sometimes at the expense of the solution quality.

In addition to those experiments, we defined the similarity between two solutions by considering either their values in the objective space or in the decision space. The purpose is to remove identical solutions. Better results are obtained if only the objective values are compared.

\end{document}